\documentclass[review,3p]{elsarticle}     

\usepackage{graphicx}
\usepackage{caption}
\usepackage{subfigure}
\usepackage{subfloat}
\usepackage{stfloats}  
\usepackage{float}
\usepackage{booktabs}
\usepackage{multirow}
\usepackage{amsmath,amssymb,latexsym}
\usepackage{lineno,hyperref}

\begin{document}

\begin{frontmatter}

\title{Hybrid attention network based on progressive embedding scale-context for crowd counting\tnoteref{mytitlenote}}




\author[1,2]{ Fusen Wang }
\author[1,2]{ Jun Sang\corref{cor1} }
\ead{ jsang@cqu.edu.cn }
\cortext[cor1]{Corresponding author}
\author[1,2]{ Zhongyuan Wu }
\author[1,2]{ Qi Liu}
\author[3]{ Nong Sang}
\address[1]{  Key Laboratory of Dependable Service Computing in Cyber Physical Society of Ministry of Education, Chongqing University, Chongqing 400044, China }
\address[2]{School of Big Data \& Software Engineering, Chongqing University, Chongqing 401331, China}
\address[3]{ School of Artificial Intelligence and Automation, Huazhong University of Science and Technology, Wuhan 430000 , China}

\begin{abstract}
The existing crowd counting methods usually adopted attention mechanism to tackle background noise, or applied multi-level features or multi-scales context fusion to tackle scale variation. 
However, these approaches deal with these two problems separately. 
In this paper, we propose an Hybrid Attention Network (HAN) by employing Progressive Embedding Scale-context (PES) information, which enables the network to simultaneously suppress noise and adapt head scale variation. 
We build the hybrid attention mechanism through paralleling spatial attention and channel attention module, which makes the network to focus more on the human head area and reduce the interference of background objects. 
Besides, we embed certain scale-context to the hybrid attention along the spatial and channel dimensions for alleviating these counting errors caused by the variation of perspective and head scale. 
Finally, we propose a progressive learning strategy through cascading multiple hybrid attention modules with embedding different scale-context, which can gradually integrate different scale-context information into the current feature map from global to local. 
Ablation experiments provides that the network architecture can gradually learn multi-scale features and suppress background noise. 
Extensive experiments demonstrate that HANet obtain state-of-the-art counting performance on four mainstream datasets.
\end{abstract}

\begin{keyword}
\texttt{Crowd counting}\sep Hybrid attention\sep Progressively embedding scale-context \sep Density map estimation
\end{keyword}

\end{frontmatter}


\section{Introduction}	  
Crowd counting based on deep learning has obtained widespread attention due to its significant applications in many large conferences, sports events, public transportation, etc.
Its purpose is to count pedestrians in single image by training a density regression network under fully supervised learning.
However, since heavy scale variation and complex texture backgrounds exist in crowd images, the counting performance still remains room for improvement.

To tackle the heavy scale variation, many methods utilized multi-column networks with different convolution kernel sizes to extract multi-scales features for adapting to different head sizes \cite{MCNN,CrowdNet,SwitchCNN,DecideNet,DADNet}. 
In addition, some approaches \cite{CP-CNN}, \cite{CAN} encoded the context information of multiple receptive field sizes to handle the problems caused by the different density distribution and continuous changes of scale in the crowd image. 
On the other hand, to avoid the disturbance from background noise, some researchers employed visual attention mechanism to generate one attention weights map from the current or previous layer features through certain means for enhancing robustness to noise \cite{DADNet}, \cite{SFANet}, \cite{SCAR}, \cite{HACNN}, \cite{ADCrowdNet}.
These methods simply divided the two problems of background noise and scale variation into different categories. 
Actually, the two problems are not independent while mutually affected and restricted each other, such as the network may not correctly distinguish small heads and leaves.

Some latest approaches also attempted to apply the attention mechanism to scale variety, but they neglect the obstruction of background noise. 
Besides, these works were often supported by some auxiliary tasks and needed to balance the calculation of multiple loss. 
This requested extra supervision data and the calculation of redundant loss \cite{SAAN,MSANet,PaDNet,DensityCNN}.

Above all, we propose one end-to-end hybrid attention network (HAN) of progressive embedding scale-context (PES) information to simultaneously suppress noise and adapt to head scale variation. 
This approach cascades several scale-context hybrid attention modules, which is composed of two parallel spatial attention and channel attention modules integrated with different head scale variation information from global to local.

To summarize, the main contributions of our work are outlined as follows:
\begin{itemize}
	\item[(1)] We propose an end-to-end network by cascading multiple hybrid attention modules, which are composed of spatial attention and channel attention in parallel.
	\item[(2)] Context information of different scales is embedded in the proposed hybrid attention module through a progressive learning strategy, which can gradually adapt to people scale variety and suppress background noise.
	\item[(3)] The results demonstrate the proposed HANet achieves the state-of-the-art performance on the four benchmark datasets.
\end{itemize}


\section{Related Work}
In this section, we briefly review some mainstream related works on CNN-based crowd counting methods. 
These mainly include crowd counting approaches based on multi-column architecture and attention mechanism.
\subsection{Multi-column methods in crowd counting}
Due to large perspective changes and varying resolution, the head size of pedestrians varies significantly in different positions of the crowd images.
Many early successful works usually conducted multi-column network architecture to address this problem.
For example, Zhang et al. \cite{MCNN} proposed Multi-Column Convolution Neural Network (MCNN) which was composed of three branches with different receptive field, aiming to tackle crowds with different densities.
Following the above work, Sam et al. \cite{SwitchCNN} trained a density classifier to adaptively select optimal regressor from three branch for different image patch (Switch CNN).
Sindagi et al. \cite{CP-CNN} added two columns of pyramid branch on the basis of MCNN to extract global and local context information for accurate counting and generating high quality density map (CP-CNN).
Boominathan et al. \cite{CrowdNet} employed two branches network with the same receptive filed of different depth, in which the shallow branch extracted small-scale features while deep branch was the opposite (CrowdNet).
However, these methods brought a large amount of computation burden and were trained slowly to improve performance.

Therefore, some latest methods attempted new means to settle scale changes.
Guo et al. \cite{DADNet} proposed multi-column dilated convolution network to capture different receptive fields required shifty head size without increasing redundant calculations.
Liu et al. \cite{CAN} proposed an end-to-end architecture that integrated multi-scale context information into current feature map in parallel (CAN).

\subsection{Attention mechanism in crowd counting}
Recently, attention mechanism has been widely incorporated to many tasks, such as object detection \cite{Detection}, image classification \cite{segmentation} and etc.
It also works for crowd counting.
The traditional attention-based crowd counting method usually generated a weight map by activation function from the current or previous layer features to mask background noise.
For instance, Zhu et al. \cite{SFANet} proposed a dual path scale fusion network, where one path was employed to generate attention map for suppressing noise, while the other path multiplied the extracted features with the attention map to generate high quality density map (SFANet).
Sindagi et al. \cite{HACNN} proposed a hierarchical attention-based network, which consisted of a spatial attention module and several global attention modules (HACNN).

Besides, some crowd counting methods of attention-based was aiming to adapt to the people head scale variety \cite{SAAN,MSANet,PaDNet,DensityCNN}.
Hossain et al. \cite{SAAN} proposed a scale-context attention network to fucus on people density variation by adding two auxiliary branch namely Global scale attention and Local scale attention (SAAN).
Jiang et al. \cite{DensityCNN} proposed multi-task density-aware network, which jointly trained density-level classification and density map estimation.
The density-level classification utilized high-level semantic information to guide density map estimation (DensityCNN).

\section{Proposed Approaches}
\subsection{Overview of network architecture}
To make the model better mask background noise on the basis of adapting to people scale variation, we design a new attention network of progressively embedding scale-context, namely HANet.
In this section, we firstly illustrate the overview architecture of our proposed HANet and then introduce each component in detail.
HANet includes three modules: (1) backbone; (2) hybrid attention; (3) backend, as shown in Figure 1.

\begin{figure*}[!t]
	\centering
	\includegraphics[width=16.0cm]{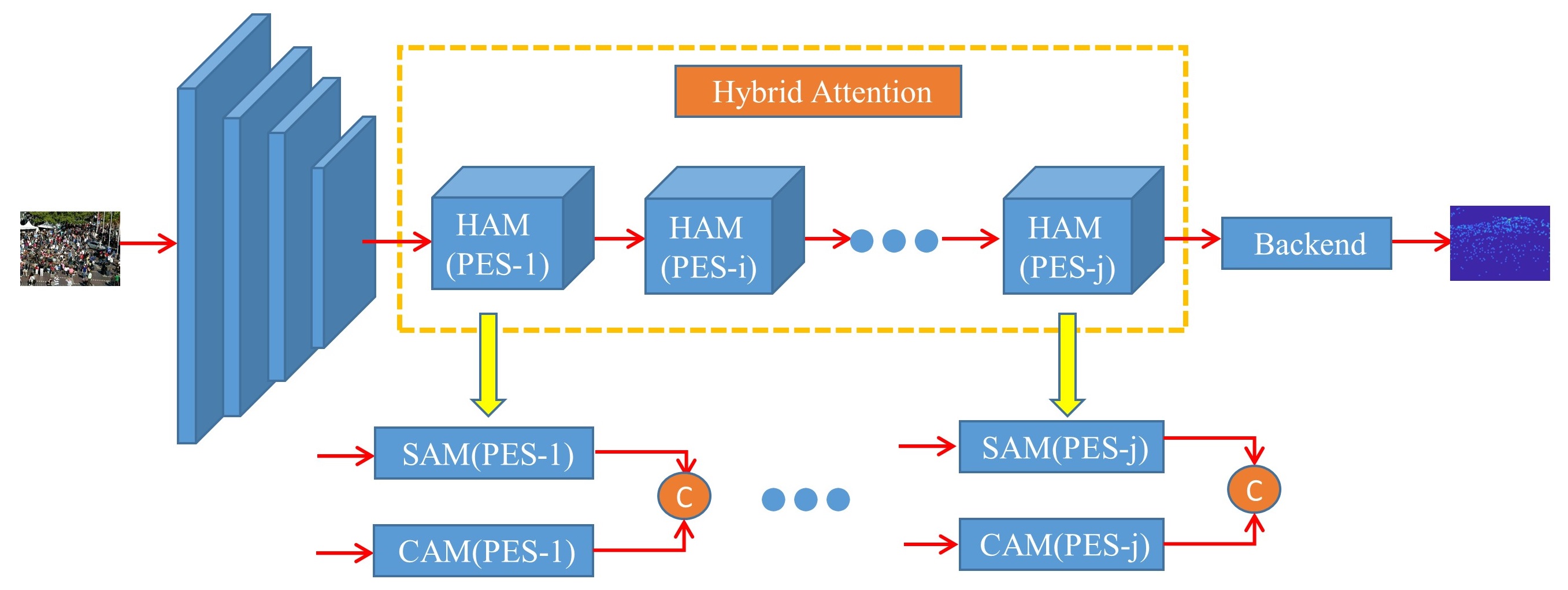}
	\caption{Overview of the proposed HANet architecture. It consists of Backbone (VGG16-BN first ten layers), Hybrid Attention (HA) that includes several cascade hybrid attention modules of progressively embedding scale-context (HAM-PES-K), and Backend.}
	\label{fig.1}
\end{figure*}

\subsection{Backbone module}
VGG16 \cite{VGG} has achieved excellent performance in crowd counting and other computer vision tasks such as object detection, classification, etc.
As W-net \cite{Wnet} indicated that VGG-BN is considered as the relatively better baseline for crowd counting task, our backbone module adopts the first ten layers of the pretrained VGG16-BN from ImageNet as front-end network to extract rich features.
Besides, since pooling layer results in losing vast location details and spatial information, we remove the final two pooling layers, which can ensure that the output of the network is 1/8 of the original image resolution.

\subsection{Hybrid attention}
In this module, we introduce the proposed hybrid attention (HA), which includes several cascade hybrid attention modules (HAM) of progressively embedding scale-context (PES) information with different scales. 
As illustrated in Figure 1, the input of the first hybrid attention module (HAM) is the backbone’s output with the size of C x H x W, and the input of the subsequent HAM is the output of the previous HAM.
We design the HAM similar with DANet \cite{DANet} through paralleling spatial attention (SAM) and channel attention (CAM) streams, except that our modules are designed with progressively embedding scale-context (PES) information in multi-scales rather than non-local mechanism.
In other words, the multi-scales information is separately integrated into several hybrid attention modules along with the spatial and channel dimensions.
The process can be formulated as follows:

\begin{equation}
{x_i} = \left\{ {\begin{array}{*{20}{c}}
		{{{\cal F}_{vgg}}(X), \quad for \; i = 0}\\
		{{\cal F}[SAM_{PES(i)}^{{\theta _i}},{\rm{ C}}AM_{PES(i)}^{{\phi _i}}]({x_{i - 1}}), \quad for \; i > 0}
\end{array}} \right.
\end{equation}

Where given an image ${X}$, ${x_0}$ denotes the Hybrid attention’s input from the output of the first ten layers of a pretrained VGG16-BN network;
${x_i}$ $(i>0)$ denotes hybrid attention module of progressively embedding scale-context implemented by function ${\cal F}$, which represents the channel-wise concatenation operation for combining ${SAM_{PES(i)}^{{\theta _i}}}$ and ${CAM_{PES(i)}^{{\phi _i}}}$; 
the output ${x_i}$ of the previous HAM is the input of the next HAM ${x_{i + 1}}$; ${x_n}$ is the final result of entire Hybrid attention.

In CANet’s method \cite{CAN}, it simply concatenates context information of different scales with current features, which ignores that the scale variations are continuous and smooth, resulting in the network to fall into a partial solution of a certain scale.
Different from CANet, our PES mechanism gradually embeds contextual information of different scales into several cascade hybrid attention modules from global to local.
The HAM’s components SAM and CAM are depicted in detail in Figure 2.

\begin{figure*}[ht]
	\centering
	\includegraphics[width=16.0cm]{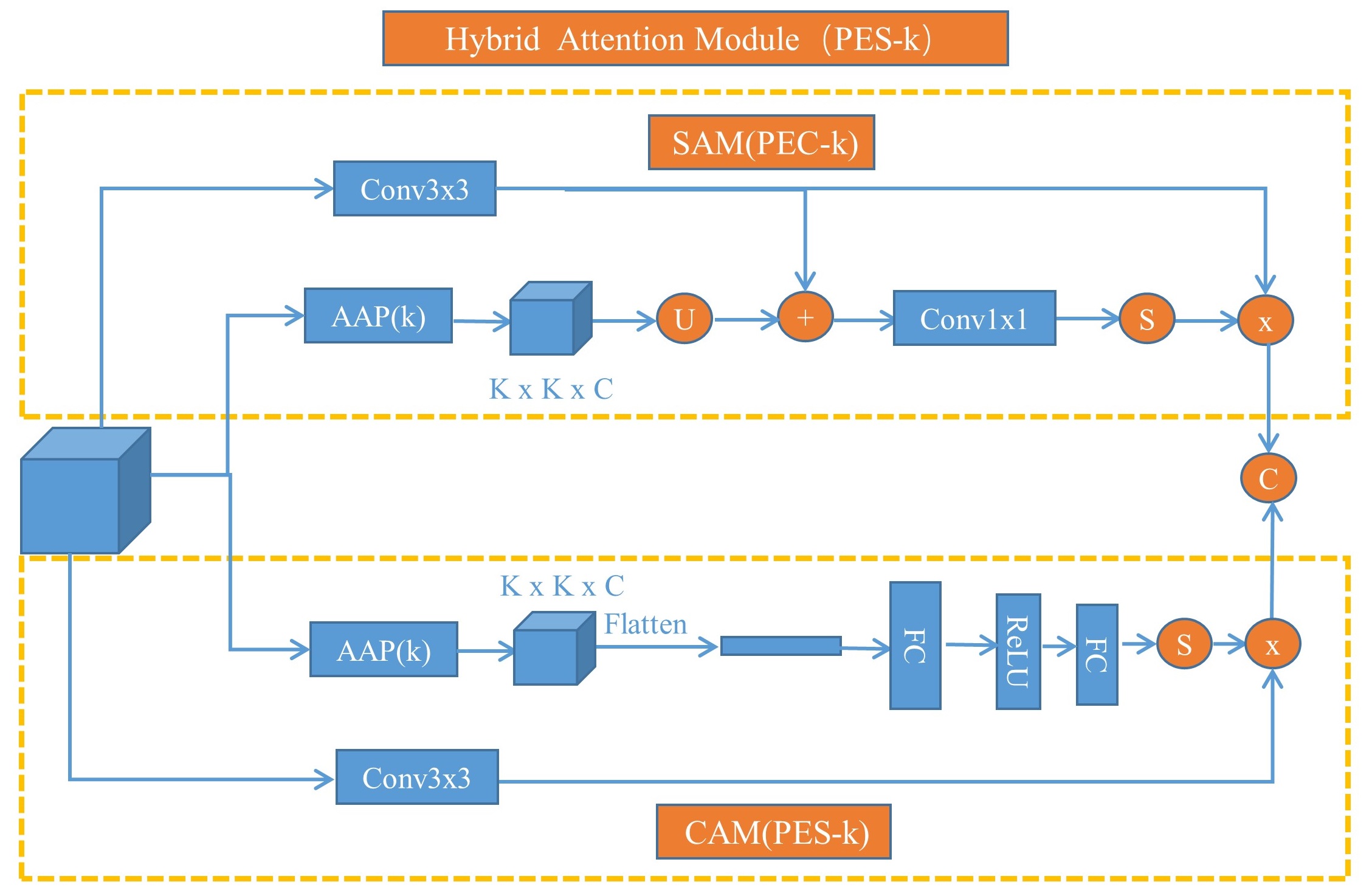}
	\caption{The detailed architecture of the Hybrid Attention Module of Progressively Embedding Scale-Context (PES-K).
		It includes two components, Spital Attention Module (PES-K) and Channel Attention Module (PES-K).
		AAP(k) indicates the adaptive average pooling of features into $K$x$K$x$C$ blocks;
		Conv $R$x$R$, Flatten, and FC represent convolution with size of $R$x$R$, flatten feature blocks of size $K$x$K$x$C$ into a $1$x$1$x${K^2}C$ vector and fully connection layer, respectively;
		U, S, ReLU, C denotes severally bilinear interpolation, sigmoid and relu activation function, and channel-wise concatenation operation.
	}
	\label{fig.2}
\end{figure*}

\subsubsection{Spatial attention module}
Different from the previous scheme \cite{CBAM}, \cite{RANet_seg}, our spatial attention module (SAM) not only encodes the probability value of the head appearing in each position of the crowd image, but also incorporates certain scale-context information into each spatial attention module (SAM).
The architecture of SAM(PES-K) is shown detailly at the top of Figure 2.
The output of SAM is defined by the following equation:

\begin{equation}
	SA{M_{PES(K)}} = {G_3}(x,\theta )*\sigma [{G_1}({G_3}(x,\theta ) + U({A_K}(x))]
\end{equation}

where for input $x$ with the size of $C$ x $H$ x $W$, it is fed into adaptive average pooling layer of scale $K$ (${A_K}$) to obtain $C$ x $K$ x $K$ scale-context blocks;
$U$ is the up-sample operation with bilinear interpolation to restore the scale-context blocks to the size of input $x$;
then add residual connection to the result of the Up step by ${G_3}$ operation, which represents a convolution layer of kernel size of 3x3;
${G_1}$ denotes a convolutional layer of kernel size of 1x1 for channel dimension reduction;
Finally, a weight map generated by Sigmoid function ($\sigma$) from the previous output is multiplied with the residual features ${G_3}(x,\theta )$.

\subsubsection{Channel attention module}
The previous channel attention attempts to strengthen important channels (foreground) and suppress the unnecessary ones (background) \cite{SENet}.
However, the lack of scale-context information limits the feature expression ability of the module.
For example, it is difficult to distinguish small heads from leaves.
We embed a scale-context blocks in CAM that is similar to SAM.
The specific architecture of CAM(PES-K) is depicted at the bottom of Figure 2, and its mathematical expression formula is as follows:

\begin{equation}
	CA{M_{PES(K)}} = {G_3}(x,\theta )*\sigma [{F_C}({F_L}({A_K}(x))]
\end{equation}

where the main operations are similar to the above SAM except $F_C$,$F_L$;
$F_L$ expresses that the $C$x$K$x$K$ scale-context blocks from $A_K$ stream are flattened into vectors with a size of $1$x$1$x${K^2}C$;
equivalent to \cite{SENet}, $F_C$ consists of fully connection 1, ReLU activate function, and fully connection 2 three layers, which are employed to learn across-channels interactive and dependence related information.
In the end, obtain a channel weight map by Sigmoid function multiplied with the residual features ${G_3}(x,\theta )$.

We conduct the hybrid attention module (HAM) by concatenation the above SAM and CAM of embedding K level scale-context information.
Then, we cascade several HAM through a progressive learning strategy as the overall hybrid attention architecture.
The advantage of this strategy is to allow the model to gradually learn features of different scales instead of the previous multi-scale feature concatenating. Inspired by \cite{CAN}, we get the best results by adopting K = 4 different scales with 1, 2, 3, 6 in light of the trade-off between resource cost and performance.
Furthermore, we adopt the embedding means of scale-context information from global to local, which is superior to the opposite approach. 
This will be validated in ablation studies of Sec. V.

\subsection{Backend}
The output of the above hybrid attention is fed to backend to generate the final density map through a set of convolution layers.
These architectures are designed as follows: {Conv (512,256,3)-BN-ReLU, Conv (256,128,3)-BN-ReLU, Conv (128,64,3)-BN-ReLU, Conv (64,1,1)}.
During the training phase, we leverage the pixel-level Euclidean Loss (MSE) to measure the error between estimated density map and ground truth:

\begin{equation}
	{{\cal L}_{MSE}} = \frac{1}{N}\sum\nolimits_{k = 1}^N {||D(X_k^{(i,j)};\theta ) - GT_k^{(i,j)}{\rm{||}}_{\rm{2}}^{\rm{2}}}
\end{equation}
where $N$ is the batch size of input’s images; 
${D(X_k^{(i,j)};\theta )}$ is the estimated density map at pixel location (i, j) for image $X_k$ with parameters $\theta$;
${GT_k^{(i,j)}}$ is the corresponding ground truth for image $X_k$.

\section{Implementation Details}
In this section, we first introduce the four main-stream datasets as well as the method of generating ground truth.
Sequentially, the data augmentation and training details are given respectively.

\subsection{Ground Truth Generation}
Similar to previous method \cite{MCNN}, we blur the annotations point of crowd images by Gaussian kernel ${G_{\mu ,\sigma }}$ with standard deviation $\mu$ and kernel size $\sigma$ to estimate the size of people head.
If there exist an annotated point at pixel $x_i$, it is denoted as $\delta (x - {x_i})$. 
Then produce density maps through convolving $\delta (x - {x_i})$.
Its mathematical expression formula is as follows:

\begin{equation}
	GT(x) = \sum\nolimits_{i = 1}^N {\delta (x - {x_i}) \times {G_{\mu ,\sigma }}} 
\end{equation}

where $N$ is the total number of annotated people in a crowd image.
For the datasets with high congested scenes, we utilize geometry-adaptive kernel to generate density map while the fixed Gaussian kernel is used in several dataset of relatively sparse scenes.
The method of producing the ground truth map in different datasets is illustrated in Table 1.

\begin{table}[]
	\centering
	\caption{Methods of generating ground truth map on different datasets.}
	\label{table_1}
	\begin{tabular}{|l|l|}
		\hline
		Datasets                    & Method               \\ \hline
		ShanghaiTech Part A \cite{MCNN} & Fixed: $\mu$ = 15, $\sigma$ = 4 \\
		ShanghaiTech Part B \cite{MCNN} & Fixed: $\mu$ = 15, $\sigma$ = 4 \\
		UCF-QNRF \cite{QNRF}           & Geometry-adaptive    \\
		UCF-CC-50 \cite{UCF_CC}          & Fixed: $\mu$ = 15, $\sigma$ = 4 \\ \hline
	\end{tabular}
\end{table}

\subsection{Datasets}
\subsubsection{ShanghaiTech}
The ShanghaiTech \cite{MCNN} dataset is divided into two parts: Part A and Part B, which contains 1198 images with a total of 330,165 annotated heads.
Part A includes 300 training images and 182 test images randomly downloaded from the Internet, where the resolutions of each image are greatly different.
Part B is composed of 400 training images and 316 test images taken from streets on Shanghai, the image resolutions of which is 768 $\times$ 1024.

\subsubsection{UCF-CC-50}
The UCF-CC-50 \cite{UCF_CC} is a very challenging crowd counting dataset due to its small size, in which the number of pedestrians each image ranges from 94 to 4,543 with an average number of 1,280 persons.
It only includes 50 images of significantly congested scenes with a total of 63,974 head annotations.
Since it is a small size dataset, we perform 5-fold cross validation by randomly selecting images to train and test our proposed approach.

\subsubsection{UCF-QNRF}
As the largest crowd dataset, the UCF-QNRF dataset \cite{QNRF} includes 1535 images of different scenarios from Internet with 1,251,642 annotations, among which it is divided into train and test set of 1201 and 334 images respectively.
The number of pedestrians in each image varies from 49 to 12,865.
Furthermore, the image resolutions are very large and its scale varied dramatically comparing other datasets.
It is a challenging dataset for crowd counting due to the high-count images and a wider variety of scenes.

\subsection{Data Augmentation}
In training phase, we randomly crop M image patches with size of m $\times$ m pixels from the original image to ensure our network can be multi-batch trained and promote performance at a lower time cost.
Table 2 details the configuration of M and m on different datasets.
Also, we ensure that these M image patches are different from each other for avoiding the patch overlapping.
In addition, for the diversity of training data, we change the RGB image into gray with probability of 0.2, and randomly horizontal flip the image with probability 0.5.
These means of data augmentation are employed in each iteration of the training process.

\begin{table}[]
	\centering
	\caption{The crop size and number of image patches on different datasets.}
	\label{Table 2}
	\begin{tabular}{|l|l|l|}
		\hline
		Datasets & M & m   \\ \hline
		ShanghaiPart A \cite{MCNN} & 4 & 128 \\ \hline
		ShanghaiPart B \cite{MCNN} & 4 & 256 \\ \hline
		UCF-CC-50 \cite{UCF_CC} & 8 & 128 \\ \hline
		UCF-QNRF \cite{QNRF} & 8 & 128 \\ \hline
	\end{tabular}
\end{table}

\subsection{Training details}
All experimental training and evaluation are implemented on the platform of PyTorch \cite{Pytorch} with a NVIDIA Tesla k80 GPU.
The baseline of the proposed HANet is leveraged from the first ten layers of pretrained VGG16-BN on ImageNet, and the other convolution layers are randomly initialized by Gaussian distributions with a mean 0 and standard deviation of 0.01.
We use SGD optimizer with learning rate of 1e-4 and weight decay of 5e-4 to train our model by minimizing the loss function Eq. (4).
The batch size is set to 8 or 16 according to GPU computing power and the number of iterations is set to 2000. 

\section{Experiments and Results}
In this section, we give the evaluation metrics and compare the results of our method with state-of-the-art methods.
In the end, we preform extensive ablation experiments to validate the effectiveness of our method.

\subsection{Evaluation metrics}
To evaluate the accuracy of our approach, the mean absolute error (MAE) and the mean squared error (MSE) are adopted as metrics. 
Specifically, equations are defined as:

\begin{equation}
	MAE = \frac{1}{N}\sum\nolimits_{i = 1}^N {|C_i^{ES} - C_i^{GT}|} 
\end{equation}

\begin{equation}
	MSE = \sqrt {\frac{1}{N}\sum\nolimits_{i = 1}^N {|C_i^{ES} - C_i^{GT}{|^2}} }
\end{equation}

where $N$ is the number of test images;
${C_i^{ES}}$ is the estimated count of the i-th image, which can be calculated by integrating the estimated density map;
${C_i^{GT}}$ is the corresponding ground truth map of the i-th image.

\subsection{Comparisons with state-of-the-art}
The extensive comparisons are conducted with state-of-the-art on four benchmark datasets to exhibit the effectiveness of our HANet. 
Furthermore, we also give the visual results. 

\begin{table}[]
	\centering
	\caption{Estimation errors on ShanghaiTech dataset.}
	\setlength{\tabcolsep}{7mm}
	\label{Table 3}
	\begin{tabular}{|l|ll|ll|}
		\hline
		\multicolumn{1}{|c|}{\multirow{2}{*}{Method}} & \multicolumn{2}{c|}{Part A}   & \multicolumn{2}{c|}{Part B}  \\ \cline{2-5} 
		\multicolumn{1}{|c|}{}                        & MAE           & MSE           & MAE          & MSE           \\ \hline
		MCNN \cite{MCNN}                                  & 110.2         & 173.2         & 26.4         & 41.3          \\
		SANet \cite{SANet}                                & 67.0          & 104.5         & 8.4          & 13.6          \\
		SCAR \cite{SCAR}                                  & 66.3          & 114.1         & 9.5          & 15.2          \\
		SFCN \cite{SFCN}                                 & 64.8          & 107.5         & 7.6          & 13.0          \\
		TEDNet \cite{TEDNet}                               & 64.2          & 109.1         & 8.2          & 12.8          \\
		DADNet \cite{DADNet}                                & 64.2          & 99.9          & 8.8          & 13.5          \\
		HACNN \cite{HACNN}                                & 62.9          & 94.9          & 8.1          & 13.4          \\
		CAN \cite{CAN}                                   & 62.3          & 100.0         & 7.8          & 12.2          \\
		MSANet \cite{MSANet}                               & 62.1          & 98.5          & 7.6          & 12.4          \\
		MMNet \cite{MMNet}                                & 60.8          & 99.0          & 7.6          & 11.7          \\
		RANet \cite{RANet}                                & 59.4          & 102.0         & 7.9          & 12.9          \\
		PaDNet \cite{PaDNet}                               & 59.2          & 98.1          & 8.1          & 12.2          \\
		PGCNet \cite{PGCNet}                               & 57.0          & \textbf{86.0} & 8.8          & 13.7          \\
		\textbf{HANet (ours)}                         & \textbf{54.9} & 91.2          & \textbf{6.8} & \textbf{11.5} \\ \hline
	\end{tabular}
\end{table}

\textbf{ShanghaiTech.} To evaluate the effectiveness of our HANet, we compare our model with 13 state-of-the-art approaches on the ShanghaiTech dataset.
As shown in Table 3, the proposed HANet achieve the lowest MAE of 54.9 and a comparable MSE of 91.2 on Part A.
For Part B, our model has also won two first places on MAE of 6.8 and MSE of 11.6.

\begin{table}[]
	\centering
	\caption{Estimation errors on UCF-QNRF dataset.}
	\setlength{\tabcolsep}{7mm}
	\label{Table 4}
	\begin{tabular}{|l|l|l|}
		\hline
		\multicolumn{1}{|c|}{\multirow{2}{*}{Method}} & \multicolumn{2}{l|}{UCF-QNRF}                   \\ \cline{2-3} 
		\multicolumn{1}{|c|}{}                        & MAE         & MSE                               \\ \hline
		MCNN \cite{MCNN}                                  & 277         & 426                               \\
		HACNN \cite{HACNN}                                & 118         & 180                               \\
		TEDNet \cite{TEDNet}                               & 113         & 188                               \\
		RANet \cite{RANet}                                & 111         & 190                               \\
		CAN \cite{CAN}                                   & 107         & 183                               \\
		SFCN \cite{SFCN}                                 & 102         & 
		\textbf{171} 					  \\
		HANet (ours)                                  & \textbf{98} & 179                               \\ \hline
	\end{tabular}
\end{table}

\textbf{UCF-QNRF.} We further assess the counting performance of HANet through comparison with 6 state-of-the-art methods on the UCF-QNRF dataset, which has the highest crowd density images and contain a variety of perspective effects, density changes.
The results are presented in Table 4.
Our approach obtains the lowest MAE of 98 and the MSE of 179 closed to the best ones.
This distinctly indicates that HANet has superior robustness against the density and scale variation.

\textbf{UCF-CC-50.} This is an extremely dense crowd dataset, which has the largest average number of each image. 
To evaluate the capability of HANet more precisely, we perform a 5-fold cross-validation that divides the dataset into 5parts, each including 40 training sets and 10 test sets. 
As shown in the Table 5, our proposed method obtains the lowest MAE with 195.2 and the second lowest MSE value of 268.6.

In the end, we take some samples from the test set of several crowd datasets for visual exhibition. 
As illustrated in Figure 3, the first, second and third column exhibit respectively the original images, ground truth and estimated density maps of HANet on ShanghaiTech Part A, B and UCF-QNRF dataset. 
From the figure, the results validate the performance of HANet in scenes with occluded backgrounds and severe scale variation.

\begin{figure*}[!t]
	\centering
	\includegraphics[width=14.0cm]{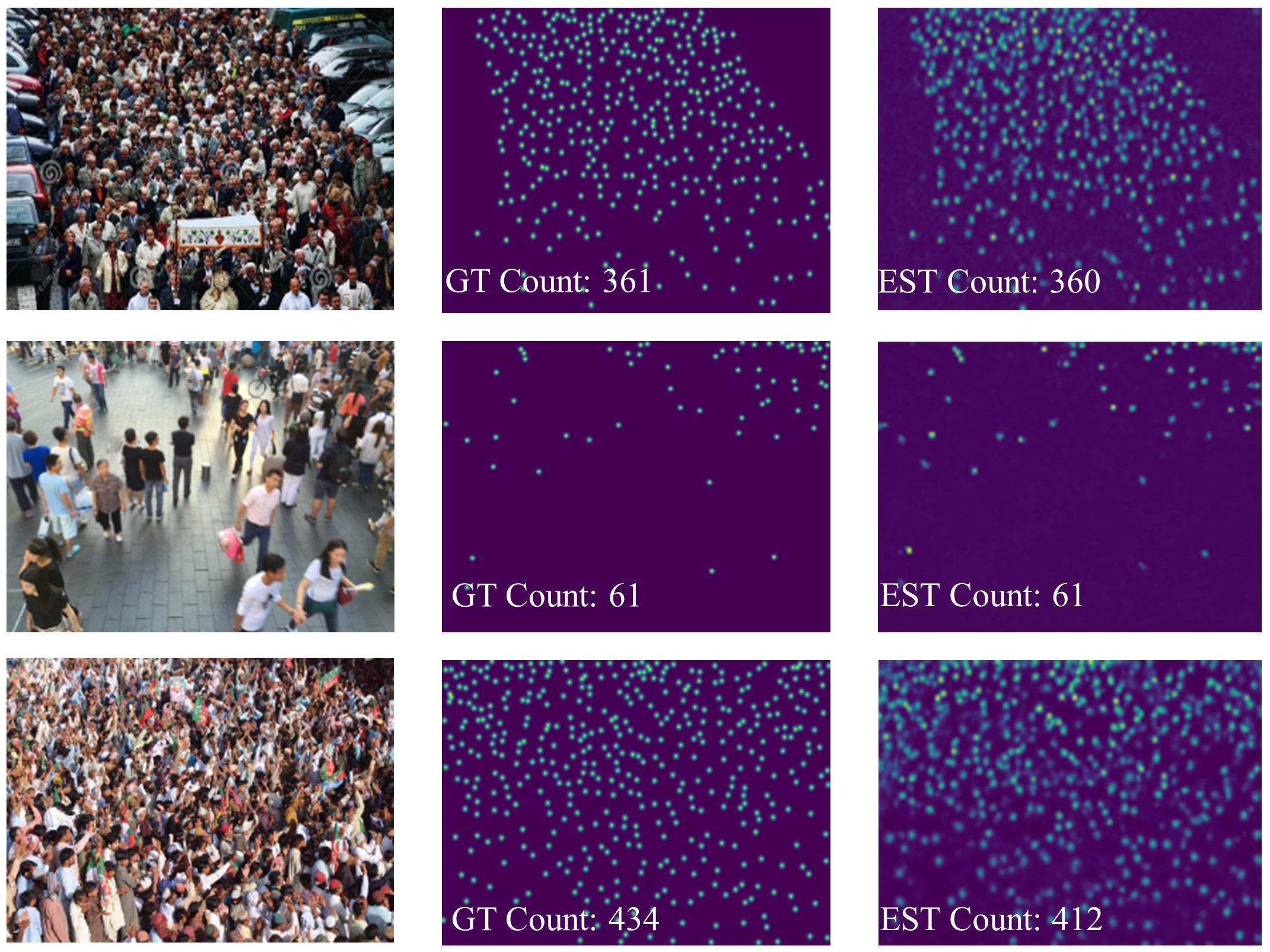}
	\caption{An illustration of estimated density maps and crowd counts generated by proposed HANet. 
		The first column shows three samples drawn from ShanghaiTech Part A, ShanghaiTech Part B and UCF-QNRF datasets. 
		The second column shows the corresponding ground truth maps. 
		The third column shows the density maps estimated by our HANet.
	}
	\label{fig.3}
\end{figure*}

\begin{table}[]
	\centering
	\caption{Estimation errors on UCF-CC-50 DATASET.}
	\setlength{\tabcolsep}{7mm}
	\label{Table 5}
	\begin{tabular}{|l|l|l|}
		\hline
		\multicolumn{1}{|c|}{\multirow{2}{*}{Method}} & \multicolumn{2}{l|}{UCF-CC-50}              \\ \cline{2-3} 
		\multicolumn{1}{|c|}{}                        & MAE            & MSE                        \\ \hline
		MCNN \cite{MCNN}                                  & 377.6          & 509.1                      \\
		SANet \cite{SANet}                                & 258.4          & 334.9                      \\
		TEDNet \cite{TEDNet}                               & 249.4          & 354.5                      \\
		CAN \cite{CAN}                                   & 212.2          & \textbf{243.7}             \\
		SFCN \cite{SFCN}                                 & 214.2          & 
		318.2                      \\
		MSANet \cite{MSANet}                               & 238.2          & 310.8                      \\
		HANet (ours)                                  & \textbf{195.2} & 268.6                      \\ \hline
	\end{tabular}
\end{table}

\subsection{Ablation Experiments}
In this section, we conduct multigroup ablation experiments to evaluate the effectiveness of different components and strategies on ShanghaiTech A, B and UCF-CC-50 datasets. 

\subsubsection{Ablation analysis on model’s components}
In this subsection, we demonstrate the performance of each module of our HANet on ShanghaiTech A dataset. 
Table 6 gives six different setting of network’s module combination to verify the effect.

\textbf{VGG16-BN first ten layers:}the backbone of our model. We load the pretrained parameters on ImageNet to VGG16-BN. 
A 1x1 convolution kernel is immediately followed to generate the predicted density map;

\textbf{VGG16-BN + Backend:} The backend includes four convolution layers (Conv512-256-3, Conv256-128-3, Conv128-64-3, Conv64-1-1), which are added to the end of backbone to regress predicted density maps more accurately;

\textbf{VGG16-BN + HAM(PES-I) + … + HAM(PES-J) + Backend(HANet):} Progressive add every hybrid attention module of embedding certain scale-context information to the end of backbone from global to local. 
Then Backend is employed to generate the estimated density map.

\begin{table*}[]
	\centering
	\caption{Estimation errors for different components of the proposed method on ShanghaiTech A dataset.}
	\label{Table 6}
	\begin{tabular}{|l|l|l|c|c|}
		\hline
		Module                                & MAE  & MSE   & \multicolumn{1}{l|}{Parameters(M)} & \multicolumn{1}{l|}{Flops(G)} \\ \hline
		VGG16-BN                              & 68.3 & 108.7 & —                                  & —                             \\
		VGG16-BN+Backend                      & 65.4 & 102.2 & —                                  & —                             \\
		VGG16-BN + HAM(PES-1) + Backend       & 61.0 & 97.4  & 11.78                              & 5.57                          \\
		VGG16-BN + HAM(PES-1,2) + Backend     & 57.9 & 91.5  & 14.56                              & 6.17                          \\
		VGG16-BN + HAM(PES-1,2,3) + Backend   & 56.6 & 95.1  & 17.68                              & 6.78                          \\
		VGG16-BN + HAM(PES-1,2,3,6) + Backend & \textbf{54.9} & \textbf{91.2}  & 22.56                              & 7.45                          \\ \hline
	\end{tabular}
\end{table*}

From the table, considering the number of parameters and calculation, the performance of our model gradually improves to 54.9/91.2 of MAE/MSE through continuously adding HAM(PES-K) to HAM(PES-1,2,3,6). 
This demonstrates the effectiveness of our approach.

\begin{figure}[ht]
	\centering
	\includegraphics[width=10.0cm]{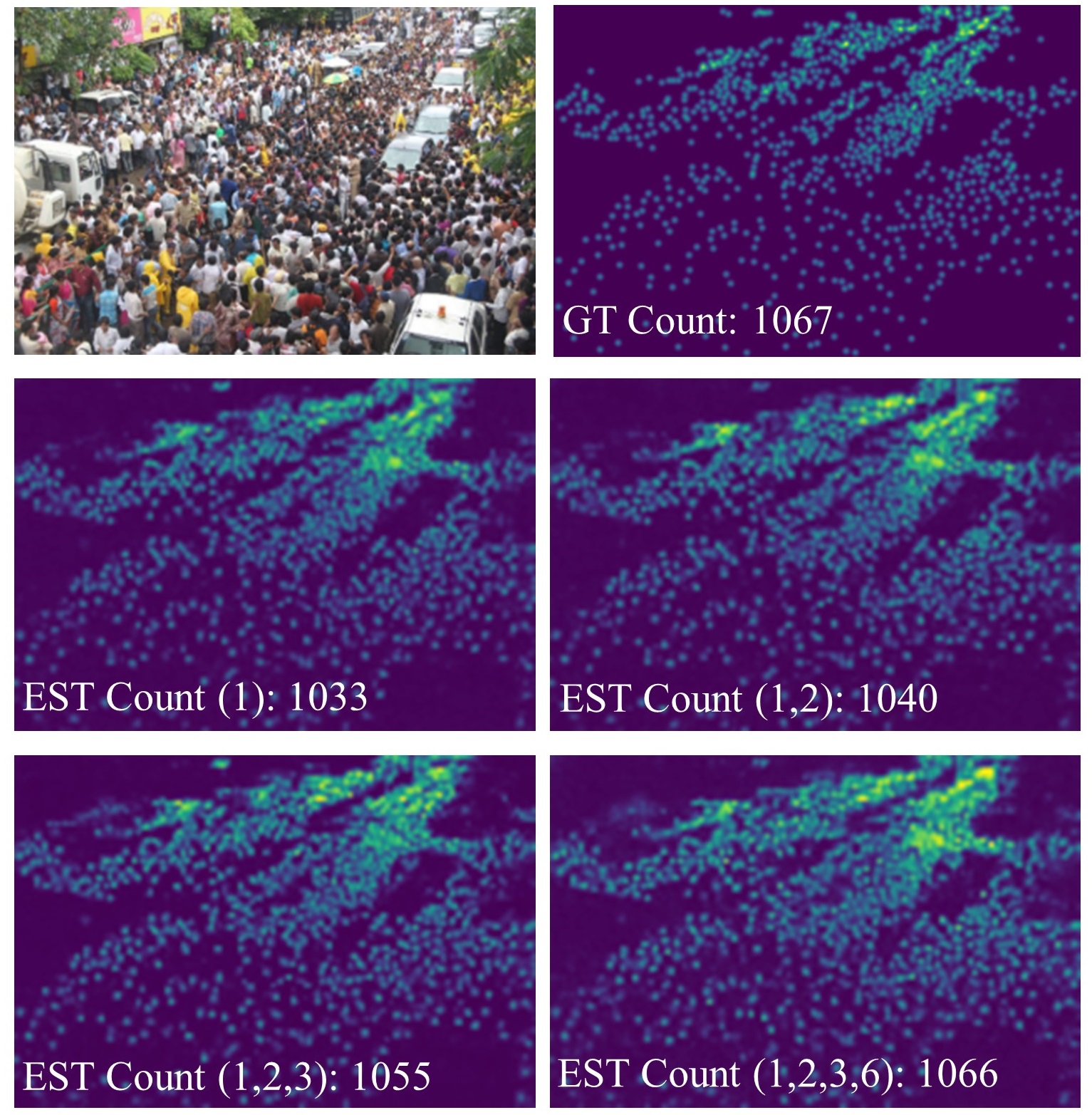}
	\caption{Estimated density maps generated different components of the model in ShanghaiTech A dataset for comparison the effectiveness of each module.
	}
	\label{fig.4}
\end{figure}

In Figure 4, we visualize the estimated density map generated by different components of the proposed model in ShanghaiTech A dataset. 
The first row represents the original image and the corresponding ground truth, respectively.
The second-third rows separately denote the estimated density map produced by multiple hybrid attention of progressively embedding scale-context information, i.e., EST Count (1,2,3,6) represents the number of people estimated by the scale combination as 1,2,3,6 in hybrid attention of progressively embedding scale-context, and the rest is similar.
From the figure, we find that the counting performance has been continuously improved close to ground truth through gradually adding attention module of embedding scale-context.

\subsubsection{Ablation analysis on HAM(PES)'s fusion methods}
Table 7 displays the comparison of two fusion methods for hybrid attention module of progressively embedding scale-context on ShanghaiTech A dataset, which represent the approaches of local to global (i.e., PES 3,2,1 or PES 6,3,2,1) and global to local (i.e., PES 1,2,3 or PES 1,2,3,6). 
We find that the model prefers to greatly the scale-learning means of the latter. From the results in the first and second rows in the Table VII, the fusion approach of PES 1,2,3 achieves better performance of MAE 56.6 and MSE 95.1. 
Similarly, in the third and fourth rows in the Table 7, the MAE and MSE of fusion method of PES 1,2,3,6 can be reduced from 56.6 and 93.2 to 54.9 and 91.2.

\begin{table}[]
	\centering
	\caption{Estimation errors of the fusion means on ShanghaiTech A dataset.}
	\setlength{\tabcolsep}{7mm}
	\label{Table 7}
	\begin{tabular}{|l|l|l|}
		\hline
		Module                                & MAE  & MSE  \\ \hline
		VGG16-BN + HAM(PES-3,2,1) + Backend   & 57.8 & 95.3 \\
		VGG16-BN + HAM(PES-1,2,3) + Backend   & 56.6 & 95.1 \\
		VGG16-BN + HAM(PES-6,3,2,1) + Backend & 56.6 & 93.2 \\
		VGG16-BN + HAM(PES-1,2,3,6) + Backend & \textbf{54.9} & \textbf{91.2} \\ \hline
	\end{tabular}
\end{table}

\subsubsection{Ablation analysis on size of image patches.}
Due to the progressively embedding scale-context (PES) of hybrid attention module, the size of the image patch also seriously affects the counting performance of the model. 
For datasets with different population area densities, the size of image patches that can improve the best performance of the network are also different. 
As shown in Table 8, we compare the impact of different image patch’s sizes for model performance on ShanghaiTech B and UCF-CC-50 datasets. 
This also demonstrates the experimental configuration in Table 2.

\begin{table}[]
	\centering
	\caption{The effects of different patch sizes with HAM(PES) module}
	\setlength{\tabcolsep}{7mm}
	\label{Table 8}
	\begin{tabular}{|l|ll|ll|}
		\hline
		\multirow{2}{*}{Size} & \multicolumn{2}{l|}{Part B} & \multicolumn{2}{l|}{UCF-CC-50} \\ \cline{2-5} 
		& MAE          & MSE          & MAE            & MSE           \\ \hline
		128x128               & 7.2          & 13.6         & \textbf{195.2}          & \textbf{268.6}         \\
		192x192               & 7.0          & 12.5         & 211.8          & 302.1         \\
		256x256               & \textbf{6.8}          & \textbf{11.5}         & 201.8          & 277.6         \\ \hline
	\end{tabular}
\end{table}

From the above table, we observed that the size of image patch has a significant influence for experimental results on the datasets of different population density distribution. 
Due to the huger number of heads and the wider crowd density on UCF-CC-50 dataset, which represent the smaller image patches usually have contained more rich heads feature information with a similar scale. 
Therefore, the patch of 128x128 (MAE of 195.2 and MSE of 268.6) achieved better performance than the larger patch of 192x 192 (MAE of 211.8 and MSE of 302.1) and 256x256 (MAE of 201.8 and MSE of 277.6). 
However, since ShanghaiTech B includes a relatively small population and a lower population density, this requires larger image patches to capture more available people scale-context information to improve the effect of model. 
In our experiments, the patches of 256x256 obtain lower MAE of 6.8 and MSE of 11.5 than patches of 128x128 (MAE of 7.2 and MSE of 13.6) and 192x192 (MAE of 7.0 and MSE of 12.5).

\section{Conclusion}
In this paper, we propose a hybrid attention network based on progressive embedded scale-context information for crowd counting. 
To distinguish human head and background more accurately, we gradually embed certain scale-context information to attention mechanism, which enables the network suppress noise and adapt to head scale changes. 
Our method is validated on four mainstream datasets, and superior counting performance is obtained in comparison with current state-of-the-art approaches.

\section*{Acknowledgment}
This work was supported by National Natural Science Foundation of China (No. 61971073).

\bibliographystyle{elsarticle-num}
\bibliography{mybibfile}

\end{document}